\documentclass[10pt,twocolumn,letterpaper]{article}

\usepackage[pagenumbers]{cvpr}

\usepackage{comment}
\usepackage[accsupp]{axessibility}  

\newcommand{\ykA}[1]{#1}

\usepackage{amsmath,amsfonts,bm,mathtools}

\definecolor{gold}{RGB}{221, 196, 65}
\definecolor{silver}{RGB}{215, 215, 215}
\definecolor{bronze}{RGB}{126, 66, 5}
\definecolor{lightgreen}{HTML}{008000}
\definecolor{lightred}{HTML}{D10000}

\def\eqref#1{equation~\ref{#1}}

\def\1{\bm{1}}

\DeclareMathAlphabet{\mathsfit}{\encodingdefault}{\sfdefault}{m}{sl}
\SetMathAlphabet{\mathsfit}{bold}{\encodingdefault}{\sfdefault}{bx}{n}

\def\@onedot{\ifx\@let@token.\else.\null\fi\xspace}

\definecolor{cvprblue}{rgb}{0.21,0.49,0.74}
\usepackage[pagebackref,breaklinks,colorlinks,allcolors=cvprblue]{hyperref}
\hypersetup{pdftitle={AvatarMix: Identity-Preserving Cross-Avatar Composition for Outfit Personalization}, pdfauthor={Zhaorong Wang, Yoshihiro Kanamori, Yuki Endo}}

\def\paperID{}
\def\confName{CVPR}
\def\confYear{2026}

\title{
AvatarMix: Identity-Preserving Cross-Avatar Composition\\for Outfit Personalization}
\author{Zhaorong Wang, Yoshihiro Kanamori, Yuki Endo\\
University of Tsukuba\\
{\tt\small zhaorong.wang1997@gmail.com, \{kanamori, endo\}@cs.tsukuba.ac.jp}\\
{\small Project page: \url{https://larsph.github.io/avatarmix/}}}

\begin{document}
\twocolumn[{
\maketitle
\begin{center}
  \includegraphics[width=\linewidth]{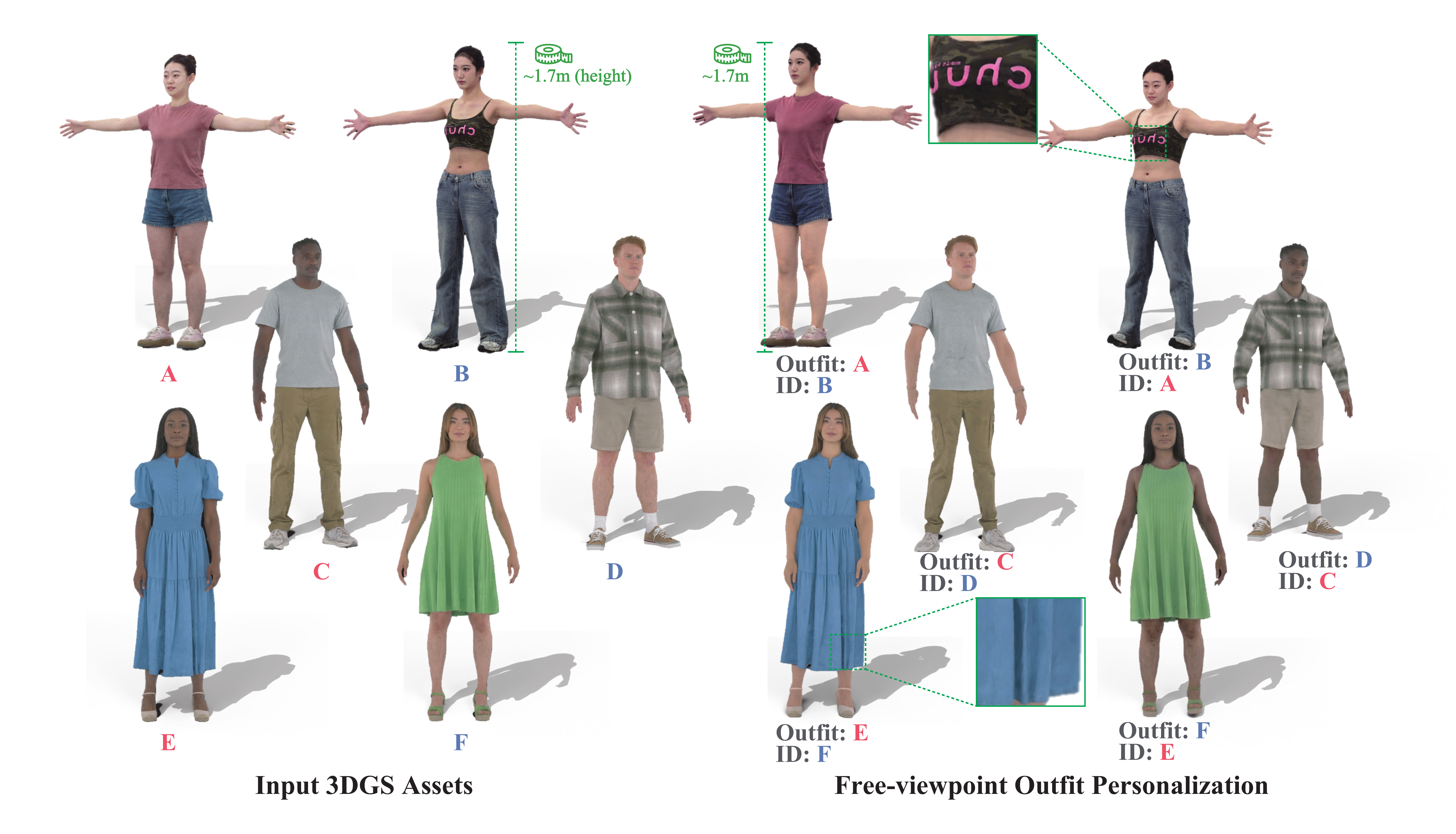}
  \captionof{figure}{AvatarMix performs free-viewpoint outfit personalization by composing a user's identity cues (head--neck, body shape and scale, and skin tone) with a model's clothed outfit in a 3DGS representation. The examples show variations in height and body proportions, cross-ethnicity skin tones, and diverse garments, while preserving fine details such as printed text and skirt folds under composition and body-shape retargeting. Examples are rendered in metric scale without post-hoc rescaling.}
  \label{fig:teaser}
\end{center}

}]
\begin{abstract}
Existing 3D avatar outfit transfer methods face distinct challenges: approaches that lift 2D edits to 3D often suffer from outfit or identity quality degradation, while those that separately model body and clothing layers are prone to intersection artifacts. We introduce AvatarMix, a \textbf{compositional paradigm} that bypasses these issues by directly composing the head and body from two high-fidelity Gaussian avatars. While this paradigm inherently preserves outfit quality and avoids intersections, it introduces challenges in creating a seamless join and maintaining appearance fidelity after body reshaping. To this end, we propose a two-tier refinement strategy: SeamFix, a localized diffusion module that refines hair and neck to ensure an artifact-free join, and an optional full-body refinement, FullbodyFix, that restores garment appearance when retargeting degrades the clothed body. Both operate on renders from an already 3D-consistent Gaussian avatar, which limits multi-view artifacts compared to 2D-to-3D lifting. To preserve the user's body identity, our mesh-based Gaussian representation enables the adaptation of a robust mesh retargeting technique, precisely reshaping the clothed body to the user's physique and robustly handling diverse body shapes. Extensive experiments demonstrate that our method achieves state-of-the-art results in outfit fidelity and identity preservation, providing a new perspective for realistic 3D outfit personalization.
\end{abstract}

\section{Introduction}
\label{sec:intro}
The ability to construct, edit, and personalize realistic 3D human avatars is increasingly crucial for immersive experiences in virtual reality, e-commerce, and digital content creation~\cite{3dhumanavatarsurvey24, 3dhumanavatarsurvey25}.
The success of 3D Gaussian Splatting (3DGS)~\cite{3dgs23} has made it more approachable to create photorealistic avatars~\cite{splattingavatar24, gauhuman24, animatablegaussians24}, providing a strong foundation for avatar editing tasks.
A key sub-task for avatar personalization is attire modification. Existing efforts can be broadly categorized by their input requirements: methods that combine a 3D avatar with 2D garment images~\cite{vton36025, gsvton24, gaussianvton24} to perform virtual try-on, and methods that transfer clothing between two existing 3D avatars~\cite{layga24, ggavatar24}.
However, both paradigms often suffer from a quality gap between the input and output avatars, with issues such as outfit fidelity degradation and geometric artifacts.
This paper focuses on a related but distinct task: \textbf{3D \ykA{avatar identity transfer for outfit personalization}},
which aims to transfer \ykA{the identity, \ie, the head, body shape, and skin tone, of a 3D Gaussian avatar representing a user who wishes to try on the outfit onto another 3D Gaussian avatar that provides the full-body garment.}
This concept has precedence in 2D image domain outfit personalization~\cite{idtransfer13}, but achieving it photorealistically in 3D with Gaussian avatars remains an open challenge.
Building upon the high-fidelity foundation enabled by 3DGS,
our work aims to enable \ykA{identity transfer for outfit personalization} between existing Gaussian avatars without compromising the quality of either the source outfit or the target identity.

Investigating the aforementioned existing paradigms reveals distinct limitations that lead to the quality gap.
1) For methods that combine a 3D avatar with 2D garment images, approaches that lift 2D edits to 3D~\cite{vton36025, gsvton24, gaussianvton24} rely on 2D generation models, often leading to texture degradation, hallucinated details, geometric inaccuracies, and multi-view inconsistencies on the edited avatar.
Among these, VTON360~\cite{vton36025} strives for fidelity by utilizing pseudo-3D pose representations and multi-view attention. But because of its reliance on a 2D generative approach without explicit 3D-space guarantees for editing, along with model generalization limitations, quality degradation on certain subjects and garments is inevitable.
While Liu \etal~\cite{tetgs25} incorporate geometric guidance for editing, their method still relies on generative texturing steps for appearance.
2) For methods that transfer clothing between two existing 3D avatars, layer-based approaches~\cite{layga24, ggavatar24} model body and clothing as separate components, introducing geometric complexity that makes them prone to intersection artifacts requiring intricate handling through geometric regularization losses and post-processing for collision. These methods also struggle to render exposed skin correctly when the body coverage of outfits differs significantly.

To address these limitations, we introduce AvatarMix,
a compositional paradigm for 3D \ykA{avatar identity transfer for outfit personalization}.
Instead of generating from 2D inputs or separating layers, we directly compose the final avatar by taking the head (for facial identity) from one high-fidelity 3DGS avatar and the clothed body (for the outfit) from another.
This explicit 3D composition inherently preserves the pristine quality (geometry and texture) of the source outfit and avoids intersection issues by design, directly addressing the core weaknesses of prior approaches.
While general 3DGS composition and editing techniques exist~\cite{tipeditor24, gaussianblock25}, AvatarMix is specifically designed to tackle the unique challenges of avatar composition: 1) creating a seamless, artifact-free join between the head and body,
and 2) adapting the clothed body geometry to \ykA{the user's original body shape},
so that the user's body identity is preserved.

AvatarMix overcomes these challenges with two key technical contributions.
First, we introduce \emph{SeamFix}, a localized diffusion refinement that targets hair and neck to ensure an artifact-free join while preserving high-fidelity face and garment details. When body reshaping degrades garment appearance, we optionally apply a full-body refinement, \emph{FullbodyFix}, on the clothed body. Crucially, both refinements are conditioned on renders of an already 3D-consistent Gaussian avatar, which constrains view inconsistency during fixing.
Second,
to adapt \ykA{the outfit to the user's original} body shape,
we strategically choose a mesh-based Gaussian representation~\cite{splattingavatar24} that enables the adaptation of a robust garment mesh retargeting method~\cite{clothfit25}. Building on this, we design \emph{GSReshape}, a mesh-based body reshaping module for Gaussian avatars that provides precise and natural deformation to match diverse physiques.

Our main contributions are:

\begin{itemize}
    \item A novel compositional paradigm for \ykA{identity transfer on high-fidelity 3D Gaussian avatars for outfit personalization}.
    \item A two-tier refinement strategy: \emph{SeamFix} for hair/neck seam correction, and an optional full-body refinement, \emph{FullbodyFix}, to restore garment appearance when body reshaping degrades quality; both operate on 3D-consistent renders to limit multi-view artifacts.
    \item Effective integration of mesh-based Gaussian representation with a robust mesh retargeting-based module, \emph{GSReshape}, for identity-preserving body shape adaptation across diverse physiques.
\end{itemize}

\section{Related Work}
\label{sec:related}

\subsection{3D Avatar Outfit Transfer and Personalization}
Editing and personalization of realistic 3D human avatars has become increasingly important across virtual reality, gaming, and digital fashion~\cite{3dhumanavatarsurvey24, 3dhumanavatarsurvey25, taoavatar25}. Avatar editing encompasses diverse tasks, including animation~\cite{animatablegaussians24, taoavatar25, vid2avatar25}, appearance and makeup modification~\cite{fenerf22, fate25, makeupprior24, avatarmakeup25}, and body reshaping~\cite{anthro17, learningsemantic23}, among which attire modification represents a critical capability for personalization.

\noindent\textbf{3D Virtual Try-On.}
A dominant approach is 3D Virtual Try-On (VTON), which applies garments onto target avatars. Traditional methods~\cite{bridson02, guan12, hahn14, lahner18, pons17} rely on physical simulation or scanning, while learning-based approaches~\cite{bhatnagar19, mir20, zhao21} employ differentiable rendering or depth-based lifting.
Recent work~\cite{dreamvton24, tech24, dreamwaltz24, dagsm24} applies Score Distillation Sampling with text-to-image diffusion models for view consistency.
Image-based 3D VTON methods leverage established 2D VTON techniques~\cite{han18, ge21, gou23, zhu23, xu24}, lifting them to 3D through multi-view editing or 2D-to-3D consistency mechanisms~\cite{vton36025, gsvton24, gaussianvton24}.
However, these VTON approaches commonly suffer from outfit quality degradation—texture loss, geometric inaccuracies, and multi-view inconsistencies—due to reliance on 2D generative priors without explicit 3D-space guarantees~\cite{vton36025}.

\noindent\textbf{Layer Modeling-based Methods.}
An alternative paradigm transfers clothing between existing 3D avatars by modeling body and garments as separate layers. Mesh-based methods~\cite{bagautdinov21, xiang21, xiang22} leverage multi-view capture, physics-based approaches~\cite{su23, li24} learn garment dynamics, while recent work employs NeRF~\cite{scarf22, delta23}, surface-based techniques~\cite{gala24}, or Gaussian representations~\cite{layga24, ggavatar24, drivable3d23}. LayGA~\cite{layga24}, for instance, uses separate Gaussian layers with geometric constraints. While enabling explicit transfer, these methods introduce geometric complexity leading to intersection artifacts requiring intricate regularization and post-processing~\cite{layga24}, and struggle with exposed skin rendering when outfit coverage varies~\cite{layga24}.

\noindent\textbf{Composition-based Methods.}
A distinct paradigm employs compositional transfer, directly combining body parts from different sources. In 2D, identity transfer methods~\cite{idtransfer13} extract and transfer user heads onto catalog model bodies, adapting body shapes through parametric image warping that manipulates semantic attributes~\cite{parametricreshaping10}. Follow-up works~\cite{replace15, imagebased17, ptvton21} improve segmentation and warping strategies. For 3D avatars, we directly compose parts from existing high-fidelity 3D Gaussian avatars, inherently avoiding quality degradation and intersection artifacts. Unlike 2D image warping, we adapt underlying 3D body geometry using mesh-based representations and 3D mesh retargeting~\cite{brouet12, guan12, ma20, clothfit25}. While general 3DGS composition exists~\cite{tipeditor24, gaussianblock25}, we specifically address avatar composition challenges—seamless joining and body shape adaptation via our GSReshape module—detailed in subsequent sections.

\subsection{Editing and Refinement of 3D Gaussians}
Recent advances enable editing and composing 3DGS scenes. Text-driven methods~\cite{gaussianeditor24wang, gaussianeditor24chen, tipeditor24, gaussctrl24} modify scenes using prompts, Score Distillation Sampling, or image-reference control. Segmentation and compositional methods~\cite{segmentany23, gaussiangrouping24, omniseg3d24, cosseggaussians24, gaussianblock25} achieve part-aware scene decomposition, enabling part-level manipulation within explicit 3DGS representations. However, these approaches focus on within-scene manipulation rather than seamlessly blending separately reconstructed avatar parts. Hybrid representation work MeGA~\cite{mega25} blends mesh faces and Gaussian hair within a single avatar using occlusion-aware rendering, but targets well-defined components from unified captures. Cross-avatar composition introduces distinct challenges—Gaussian artifacts, segmentation errors, and seam misalignment—requiring specialized refinement.
To address quality degradation issues in 3DGS scenes, refinement methods have been recently explored primarily for artifacts arising from sparse-view reconstruction and challenging novel viewpoints. Generative prior-based approaches enhance quality by refining rendered 2D views before lifting to 3D~\cite{deceptive24, enhancer24, gaussianobject24, gsfix3d25} or directly refining 3D Gaussian representations~\cite{splatformer25}. Difix3D+~\cite{difix3d25} leverages diffusion priors for progressive 3D updates and render-time refinement, but is designed for general scenes and focuses on smoothing reconstruction artifacts rather than handling avatar-specific issues such as garment seams, fine facial details, or large missing regions; even follow-up work like GSFix3D~\cite{gsfix3d25} primarily targets reconstruction defects and cannot reliably fill substantial holes. We therefore adapt Difix3D+ to cross-avatar composition by fine-tuning two avatar-specific refinement modules: SeamFix, a localized diffusion refiner for head–neck seams, and FullbodyFix, a full-body restoration module that addresses garment and skin artifacts within a unified two-tier diffusion refinement framework.

\begin{figure*}[t]
  \centering
  \includegraphics[width=\linewidth]{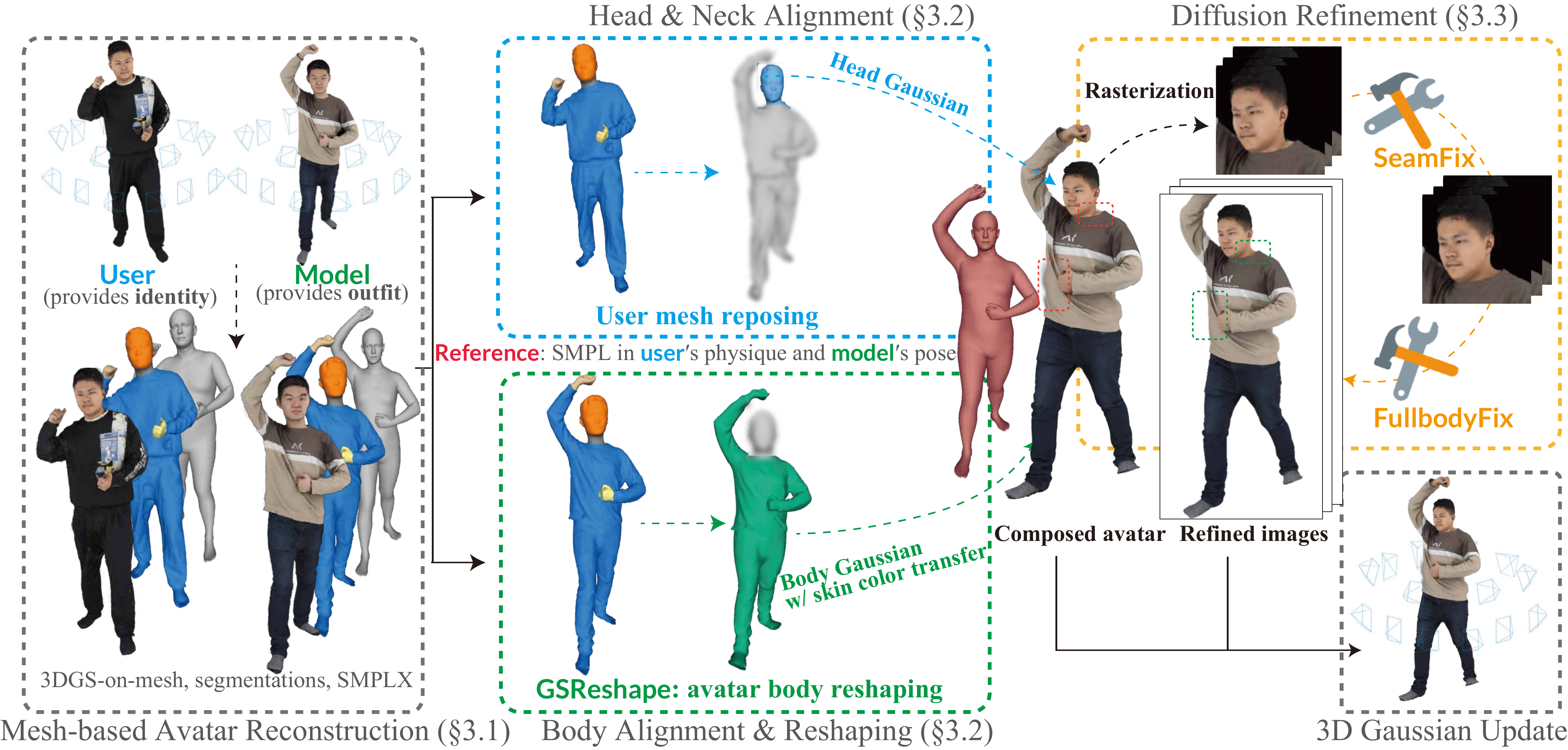}
\caption{\textbf{Overview of AvatarMix.} Given multi-view images of a \ykA{U}ser and a \ykA{M}odel,
we first \ykA{employ} Mesh-Based Avatar Reconstruction (\cref{sec:prep}) \ykA{with semantic segmentation}.
We then perform Cross-Avatar Geometric Composition (\cref{sec:comp}) by aligning the user’s head and neck to the \ykA{M}odel’s pose (Head \& Neck Alignment) and reshaping the \ykA{M}odel’s clothed body via our GSReshape module (Body Alignment \& Reshaping) so that the body geometry matches the \ykA{U}ser’s physique, yielding a composite mesh-based Gaussian avatar. Finally, our two-tier diffusion refinement (SeamFix for localized head–neck seams and optional FullbodyFix for full-body garment/skin artifacts; Sec.~\ref{sec:refine}) operates on rendered views, followed by 3D Gaussian fine-tuning,
to produce the final \ykA{identity-transfer} result.
}
   \label{fig:overview}
\end{figure*}

\begin{figure}[t]
  \centering
  \includegraphics[width=\linewidth]{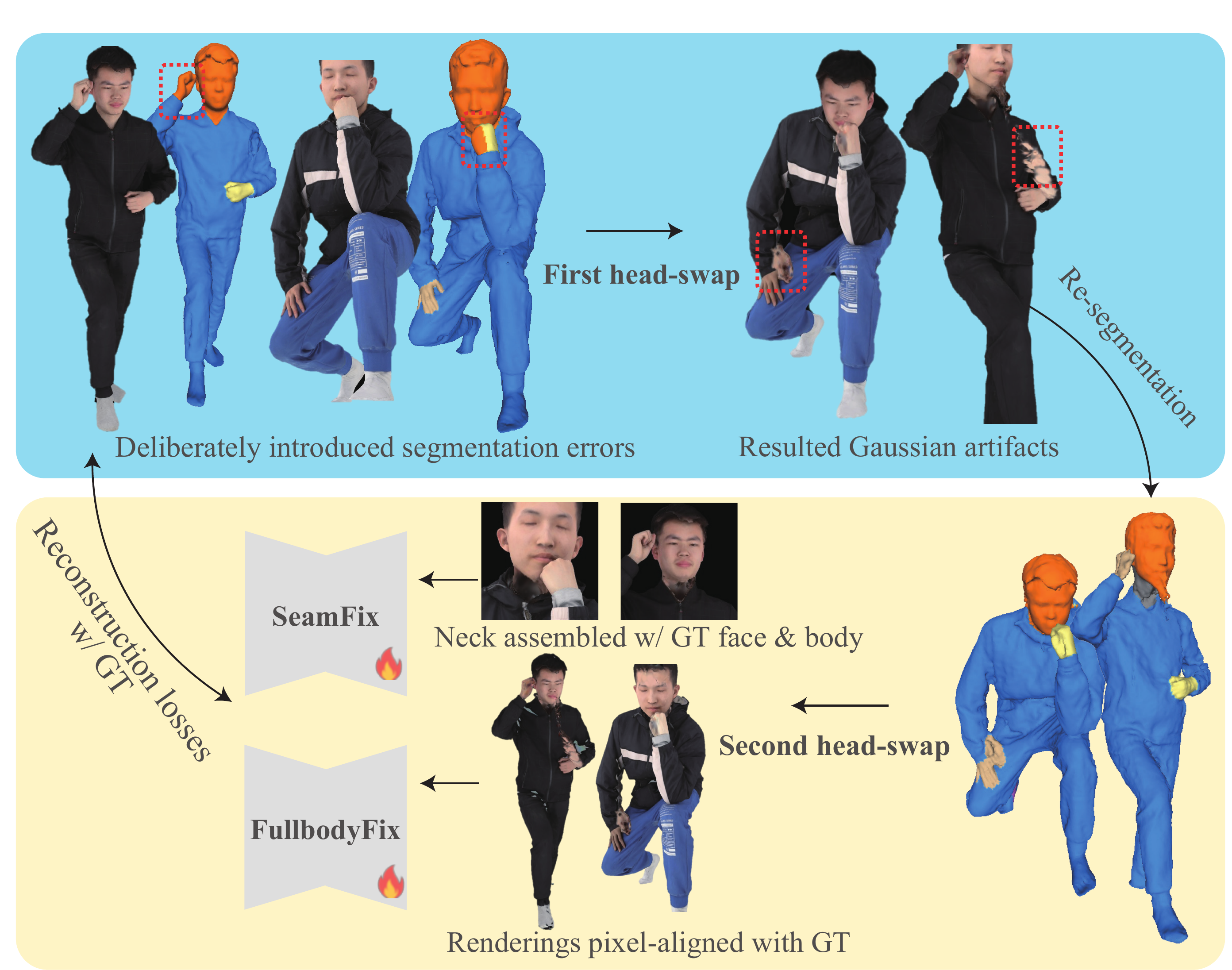}

   \caption{\textbf{Training strategy for SeamFix and FullbodyFix.} Top: starting from two avatars A and B, we deliberately introduce segmentation errors by using the original 4D-Dress with SAM voting and perform a first head-swap A$\rightarrow$B, which produces Gaussian artifacts at the head, neck, and hands. After re-segmentation and a second head-swap B$\rightarrow$A, we obtain double-swapped avatars that are pixel-aligned with the ground-truth avatars but exhibit diverse composition-induced artifacts. Bottom: using the ground-truth renderings as supervision, SeamFix is trained on portrait crops where the artifact neck from the double-swapped avatar is combined with the ground-truth face and body under 2D masks, and FullbodyFix is trained on full-body renders. Both branches fine-tune LoRA adapters of a Difix-based diffusion backbone to correct artifacts from cross-avatar composition.}
   \label{fig:training}
\end{figure}

\section{Method}
To achieve \ykA{identity transfer for outfit personalization},
\ykA{our method},
AvatarMix, introduces a compositional paradigm that directly combines parts from two existing 3D Gaussian avatars: a \textbf{User \ykA{a}vatar} $\mathcal{A}_u$
(providing \ykA{head, body shape, and skin tone})
and a \textbf{Model \ykA{a}vatar} $\mathcal{A}_m$ (providing the clothed outfit). This explicit 3D composition inherently preserves high fidelity by design.
However, successfully \ykA{personalizing outfits via identity transfer} necessitates adapting the clothed body geometry to match varying physiques.
To enable this body reshaping capability, we adopt a mesh-based Gaussian representation where 3D Gaussians $\mathcal{G}$ are constrained to a reconstructed mesh surface $\mathcal{M}$, following SplattingAvatar~\cite{splattingavatar24}. This explicit mesh structure allows us to leverage robust garment retargeting techniques~\cite{clothfit25} for precise body shape adaptation.

As illustrated in \cref{fig:overview}, our pipeline consists of three main stages: 1) \textbf{Mesh-Based Avatar Reconstruction} (\cref{sec:prep}) reconstructs high-fidelity mesh-based Gaussian avatars with semantic segmentation; 2) \textbf{Cross-Avatar Geometric Composition} (\cref{sec:comp}) extracts and aligns the user's head and neck while reshaping the model's clothed body to match the user's physique, yielding a composite avatar; 3) \textbf{Diffusion Refinement} (\cref{sec:refine}) applies diffusion-based inpainting exclusively to the head-body seam region, preserving high-fidelity outfit and face details while correcting 3D segmentation errors via robust 2D masks. Finally, we fine-tune the composite 3DGS representation using refined multi-view images as supervision.

\subsection{Mesh-Based Avatar Reconstruction}
\label{sec:prep}

Our compositional paradigm requires high-fidelity mesh-based Gaussian avatars for both the user and model. Given multi-view image sets $\{\mathcal{I}_u, \mathcal{I}_m\}$ of the two subjects, we reconstruct hybrid 3D representations that combine explicit mesh geometry with Gaussian appearance, enabling both photorealistic rendering and geometric manipulation.

\noindent\textbf{Avatar Reconstruction.} To enable robust body reshaping operations in subsequent stages, we require an explicit mesh representation. We employ NeuS2~\cite{neus2} to reconstruct a high-fidelity mesh $\mathcal{M} \in \mathbb{R}^{N_v \times 3}$ for each avatar, where $N_v$ denotes the number of vertices. NeuS2 is particularly suitable for our task as it robustly handles complex topologies common in loose or intricate clothing, producing meshes that faithfully capture garment geometry without requiring watertightness. To achieve photorealistic rendering while maintaining the explicit structure necessary for geometric operations, we adopt Splatting Avatar~\cite{splattingavatar24} to define 3D Gaussians on the reconstructed mesh surface. Specifically, each mesh vertex $\mathbf{v}_i$ is associated with a 3D Gaussian $\mathcal{G}_i$ characterized by its position (constrained to $\mathbf{v}_i$), covariance, color, and opacity. This mesh-constrained Gaussian representation provides efficient, high-quality rendering via differentiable splatting while ensuring that appearance modifications remain anchored to the underlying geometry.

\noindent\textbf{Semantic Segmentation and Pose.} To facilitate part-aware composition, we employ a modified version of 4D-Dress~\cite{4ddress} to obtain semantic segmentation of mesh vertices into distinct regions: head, torso skin, left/right arm skin, left/right leg skin, and clothing. We modify the original method by removing SAM-based voting, which improves face and hand separation accuracy by preventing the over-aggregation of skin regions. Additionally, our method requires fitted SMPL-X parameters~\cite{smplx} for each subject, which provide parametric body models to guide pose alignment and body shape adaptation in the composition stage.

\subsection{Cross-Avatar Geometric Composition}
\label{sec:comp}

This stage performs the core geometric operations to combine the user's identity with the model's outfit through cross-avatar composition.

\noindent\textbf{Head and Neck Alignment.} To preserve the user's facial identity, we extract and align their head and neck region to the model's pose. We first transfer~\cite{robusttransfer} Linear Blend Skinning (LBS) weights from the user's SMPL-X model to their high-resolution mesh $\mathcal{M}_u$ using nearest neighbor mapping. Using these weights, we repose $\mathcal{M}_u$ to match the model avatar's pose, ensuring precise head alignment. We then extract the head and neck vertices (and their associated Gaussians) based on the semantic segmentation. Notably, we include the neck to preserve as much identity information as possible; the resulting neck-garment seam will be seamlessly refined in the subsequent diffusion stage.

\noindent\textbf{GSReshape: Body Reshaping via Mesh Retargeting.} The key challenge in cross-avatar composition is adapting the model's clothed body to fit the user's physique. We employ the garment retargeting method from Huang et al.~\cite{clothfit25}, originally designed for fitting garments to body shapes. We adapt it for our body reshaping task: deforming the model's clothed body mesh $\mathcal{M}_m^{\text{body}}$ (including both clothing and exposed skin) to match the user's body shape represented by their SMPL-X mesh $\mathcal{M}_u^{\text{SMPL}}$. This Gaussian-avatar-specific adaptation of mesh retargeting constitutes our GSReshape module.
However, this adaptation introduces three technical challenges.
\textit{(a) Hand-aware skin tightness:} In the original retargeting formulation, high fit weights on skin vertices drive the mesh toward the underlying SMPL body, which is particularly problematic for articulated regions such as hands: even when the optimized mesh looks plausible, the attached Gaussians can become severely distorted due to the complex hand geometry. Using low fit weights on skin avoids this issue but produces a glove-like effect where the hands no longer follow the target body shape. We resolve this trade-off by removing the hand geometry from the SMPL-X mesh used during retargeting, thereby preventing the SDF-based barrier energy from pushing the avatar hands outward, and by employing low fit weights on hand vertices. This combination keeps the hand region clean and non-inflated in the Gaussian avatar, at the cost that hand shape adaptation is not modeled; we discuss this limitation and provide an ablation in the supplementary material.
\textit{(b) Intersection-free initialization:} The retargeting method requires an intersection-free initial state between the clothed mesh and SMPL skeleton. Since clothed avatars are more prone to skeleton intersections than pure garments (especially at limbs), we optimize the SMPL skeleton vertices using an as-rigid-as-possible (ARAP) deformation, ensuring they lie inside the clothed avatar via negative signed distance field (SDF) values while maintaining skeleton rigidity.
\textit{(c) Computational efficiency:} To reduce computational cost, we perform retargeting on simplified meshes and transfer the deformation back to the original high-resolution mesh by finding the closest surface point for each original vertex. The Gaussians attached to the mesh are deformed accordingly.

\noindent\textbf{Skin Tone Transfer.}
To harmonize skin appearance between the user's head and the model's body, we perform a global skin-tone transfer in Lab color space. We first estimate an opacity-weighted mean and variance of face colors from the user's facial Gaussians, and analogous statistics from the model's skin Gaussians. We then normalize the model's skin colors using its statistics and re-scale them to match the user's statistics before converting back to RGB. This global, opacity-aware color transfer matches the model body's skin tone to the user's while preserving local shading and high-frequency detail; implementation formulas are provided in the supplementary material.

With both components prepared, we directly combine them by replacing the head region of the reshaped model body with the aligned user head and neck, yielding a composite mesh-based Gaussian avatar $\mathcal{A}_c = \{\mathcal{M}_c, \mathcal{G}_c\}$ that preserves both the user's identity and the model's outfit geometry.

\subsection{Diffusion Refinement}
\label{sec:refine}

While cross-avatar composition yields a geometrically valid avatar, visual artifacts may persist in two key areas: seam artifacts at the head-body boundary, particularly around hair and neck regions; and garment appearance degradation resulting from body reshaping deformations. To address these issues, we employ a two-tier diffusion refinement strategy that operates on rendered images from the already 3D-consistent Gaussian avatar. By refining renders of a coherent 3D representation rather than independently editing 2D images before 3D lifting, our approach inherently constrains multi-view inconsistency.

\noindent\textbf{SeamFix: Localized Neck Refinement.}
To ensure an artifact-free join while preserving the pristine quality of unaffected regions, SeamFix applies diffusion-based refinement exclusively to the hair and neck areas. As shown in~\cref{fig:training}, we synthesize training pairs via a double-swapping procedure: starting with two avatars A and B, we first perform composition from A to B, then immediately reverse the process from B back to A. This double-swap produces realistic artifact patterns around the seam without requiring manual annotation. Importantly, both the ground truth head and body regions and the twice-swapped results are geometrically aligned after two composition operations, enabling us to construct training inputs by using the double-swapped rendered head and neck as noisy inputs while treating the ground truth as supervision targets.

To improve robustness, we incorporate an augmentation strategy that uses 2D segmentation masks~\cite{graphonomy19} extracted from the second swap rendering. These masks often contain incomplete neck regions with missing pixels, simulating real-world segmentation failures. We crop a portrait region encompassing the head and neck, then dilate the boundary to include collar context before feeding to the diffusion model. Additional implementation details are provided in the supplementary material.

\noindent\textbf{FullbodyFix: Optional Full-Body Restoration.}
When body reshaping operations degrade the appearance of the clothed body region, we optionally apply a full-body diffusion refinement. Training data for FullbodyFix uses full-body renders from double-swapped avatars, which exhibit various garment and skin artifacts due to the repeated composition process. Unlike SeamFix, the model learns to restore the complete appearance without compositing with ground truth body parts. Although FullbodyFix performs restoration over the entire human region, it operates on renders from a 3D-consistent Gaussian avatar, which mitigates multi-view inconsistency compared to methods that apply 2D try-on generation followed by 3D lifting. In our current implementation, FullbodyFix is applied manually when visual inspection indicates quality degradation; we do not employ an automatic triggering mechanism.

\noindent\textbf{Network Architecture and Training.}
We build upon the pretrained Difix model, which is trained to fix Gaussian splatting reconstruction artifacts using Stable Diffusion Turbo as its backbone. The original Difix model employs LoRA adapters to train both the full UNet and VAE decoder. In AvatarMix, we freeze all original Difix LoRA weights and introduce two new trainable LoRA adapters: one for the UNet and one for the VAE decoder. SeamFix and FullbodyFix share this architectural design but employ different LoRA ranks to balance capacity and efficiency: SeamFix uses rank-8 adapters while FullbodyFix uses rank-16 adapters. All other training configurations remain consistent across both modules unless otherwise specified.

\section{Experiments}
\label{sec:exp}

\begin{table*}[t]
\centering
\small
\caption{\textbf{Quantitative comparison on THUman2.0.} We report Editing Target DINO similarity, which measures preservation of the edited region (upper garment for VTON360; clothed body for AvatarMix); Head+Neck DINO similarity for facial identity and neck seam quality; warping-based RMSE on edited images for multi-view consistency; and user study preference percentages for Overall quality, view Consistency, and Facial quality. ``N/A'' indicates the metric is not applicable for the method.}
\setlength{\tabcolsep}{5.5pt}
\begin{tabular}{lcccrrr}
\toprule
Method & Edit. Tar. DINO $\uparrow$ & Head+Neck DINO $\uparrow$ & Warp. RMSE $\downarrow$ & Vote Overall $\uparrow$ & Vote Consist. $\uparrow$ & Vote Facial $\uparrow$ \\
\midrule
VTON360 & 0.633 & 0.786 & 0.0276 & 8.70\% & 10.43\% & 7.83\% \\
TIP-Editor & N/A & 0.356 & 0.0388 & 2.61\% & 2.61\% & 0\% \\
AvatarMix (Ours) & \textbf{0.883} & \textbf{0.818} & \textbf{0.0175} & \textbf{88.69\%} & \textbf{86.96\%} & \textbf{92.17}\% \\
\bottomrule
\end{tabular}
\label{tab:quant_results}
\end{table*}

\begin{figure*}[t]
  \centering
   \includegraphics[width=\linewidth]{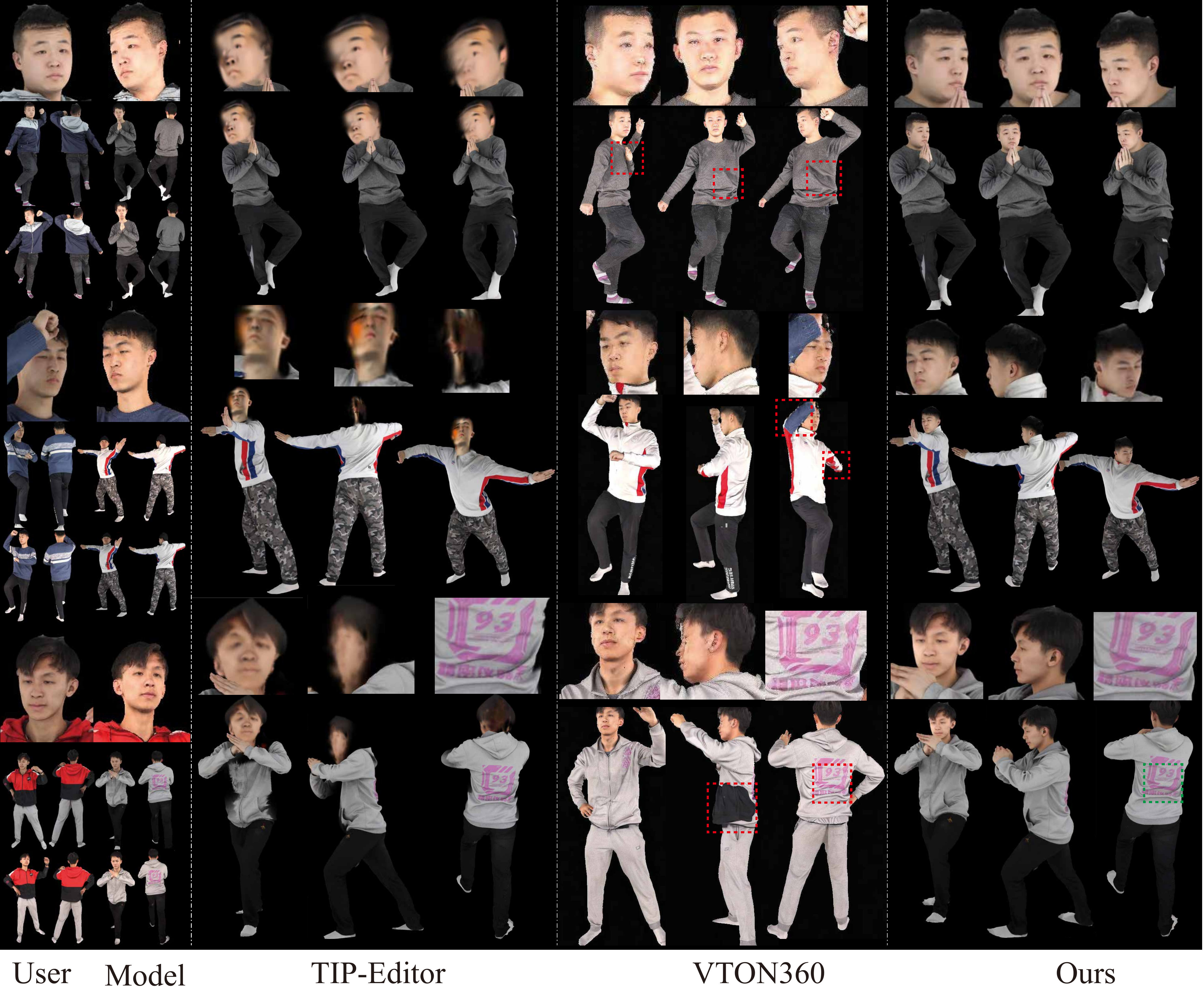}

   \caption{\textbf{Qualitative comparison with TIP-Editor and VTON360.} For each user--model pair, we show the input user and model images (front/back under two lighting conditions), followed by three-view outputs of TIP-Editor, VTON360, and AvatarMix. Zoomed insets highlight faces and garment regions, and red dashed boxes mark typical failure cases of existing methods, including view inconsistency, unnatural garment wrinkles, and degraded hands. AvatarMix better preserves facial identity, garment texture, and seam quality while avoiding these artifacts.}
   \label{fig:comparison}
\end{figure*}

\begin{figure}[t]
  \centering
  \includegraphics[width=\linewidth]{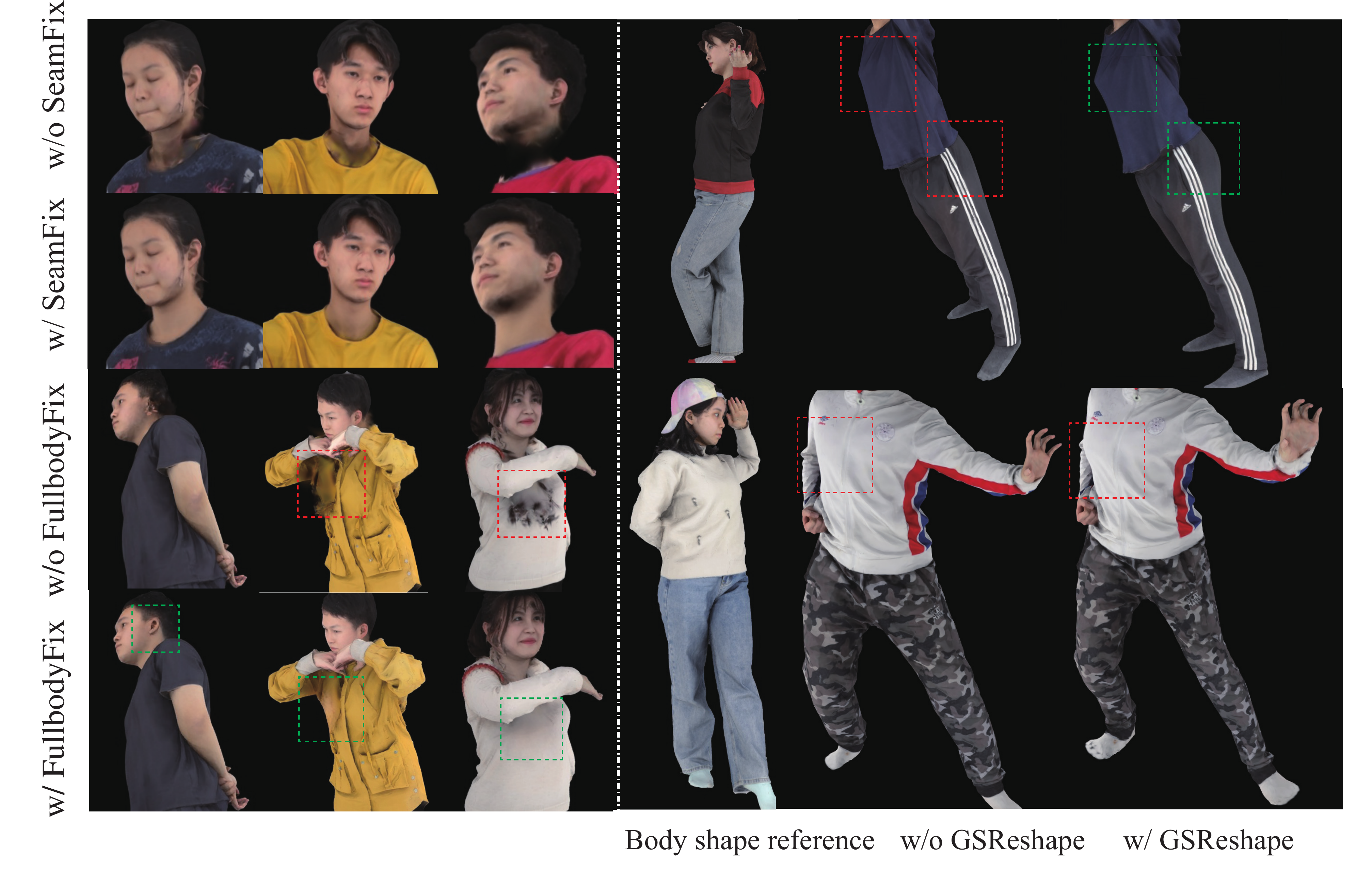}

  \caption{\textbf{Ablation of diffusion refinement and GSReshape.} Left: rows show compositions without SeamFix, with SeamFix, without FullbodyFix, and with FullbodyFix for three user--model pairs, illustrating how SeamFix cleans head--neck seams and FullbodyFix restores garment appearance while preserving face and outfit details. Right: for two subjects, we compare the reference model body, composition without GSReshape, and composition with GSReshape; heads are cropped in the latter two columns so that differences in body shape and garment fit are easier to see.}
  \label{fig:ablation}
\end{figure}

\subsection{Experimental Setup}
\noindent\textbf{Dataset.}
We conduct our evaluation on the THUman2.0 dataset~\cite{thuman221}, which contains 526 reconstructed clothed human subjects with diverse body shapes, clothing styles, and articulated poses. Following the data split of VTON360~\cite{vton36025}, we use 110 subjects as the test set and the remaining subjects for training our reconstruction and diffusion networks. On the 110 test subjects, we construct user–model pairs by randomly sampling identity/garment combinations and render 36 viewpoints per pair. For AvatarMix and VTON360, all test pairs and views are used in the quantitative evaluation, while TIP-Editor is evaluated on a subset of pairs with the same view protocol. We also design task-specific metrics and a user study tailored to identity transfer for outfit personalization.

\noindent\textbf{Baselines.}
We compare our method against two recent state-of-the-art approaches that address related but distinct avatar editing tasks. VTON360~\cite{vton36025} performs 3D virtual try-on by applying upper garments to target avatars through per-view 2D generation followed by 3D lifting. TIP-Editor~\cite{tipeditor24} enables localized 3D Gaussian splatting scene editing and can be configured for head replacement tasks. Our method, AvatarMix, performs head replacement combined with body shape adaptation through the GSReshape module, followed by SeamFix for seamless joining and optional FullbodyFix for appearance restoration when needed.

\noindent\textbf{Evaluation Metrics.}
We employ three quantitative metrics computed on per-view edited images, complemented by a user study. Editing Target DINO measures how well each method preserves the edited region (upper garment for VTON360, clothed body for AvatarMix) using DINO~\cite{dinov224} similarity between outputs and their respective references; this metric is not applicable to TIP-Editor, which edits only the head. Head and Neck DINO evaluates facial identity and seam quality using DINO similarity on a head-and-neck mask between edited results and the ground-truth user avatar. Warping-based RMSE measures multi-view consistency by aligning neighboring views via dense 2D correspondences and computing the root-mean-squared error between one view and the other warped into its coordinate frame~\cite{met3r25}. Detailed protocols and implementation choices are provided in the supplementary material.
\subsection{Implementation Details}
Both SeamFix and FullbodyFix are implemented on top of the Difix3D+ backbone~\cite{difix3d25}. Following~\cref{sec:refine}, we freeze the original Difix LoRA weights and add new LoRA adapters on the UNet and VAE decoder, using a rank-8 for SeamFix and a rank-16 for FullbodyFix, and the same rank-4 for the VAE decoder. Our GSReshape module builds on the intersection-free garment retargeting method of Huang \etal~\cite{clothfit25}; we largely follow their default hyper-parameters, but use reduced fit weights and a similarity term on hand vertices as discussed in~\cref{sec:comp}, together with the standard SDF-based skeleton regularizers from~\cite{clothfit25}. 

\subsection{Quantitative Evaluation}
\cref{tab:quant_results} summarizes the quantitative results on THUman2.0. Overall, AvatarMix outperforms competing methods on all metrics for which it is applicable. In particular, our method attains the highest Editing Target DINO and Head+Neck DINO similarities, even under stricter evaluation protocols that use all 36 views and include body reshaping and pose changes, indicating superior preservation of both garment appearance and facial identity. The lowest warping-based RMSE further confirms that our compositional and refinement pipeline produces more view-consistent edits than both the 2D-lifting-based VTON360 and the localized editing of TIP-Editor.

\noindent\textbf{User Study.}
To complement the quantitative metrics with perceptual evaluation, we conduct a user study with 23 participants on a set of comparative questions covering multiple
\ykA{identity-transfer (User–Model) cases for outfit personalization}
(15 total judgments per participant). For each case, we present multi-view image grids of VTON360, TIP-Editor, and AvatarMix, and ask participants to make forced-choice selections of the best method along three dimensions: Overall realism and visual quality, consistency of appearance across views, and preservation of facial features and seam quality. As reported in~\cref{tab:quant_results}, AvatarMix is preferred by a large majority of participants for all three criteria, while VTON360 is chosen only occasionally and TIP-Editor is rarely favored. These trends align closely with the quantitative metrics and support the effectiveness of our compositional approach in terms of both objective scores and human perception.

\subsection{Qualitative Evaluation}
We present visual comparisons among VTON360, TIP-Editor, and AvatarMix on the THUman2.0 dataset in \cref{fig:comparison}. Our qualitative results highlight three key aspects: facial identity preservation, neck seam quality achieved by SeamFix, and outfit fidelity maintained through our compositional approach and enhanced by FullbodyFix when applied. The comparisons demonstrate the effectiveness of our method in preserving high-fidelity details from both source avatars while achieving seamless integration. We further provide qualitative ablations of SeamFix and FullbodyFix, and of GSReshape in \cref{fig:ablation}. Additional visual results, including multi-view renderings and diverse body shape examples, are provided in the supplementary material.

\section{Conclusions}
 We presented AvatarMix, a compositional framework for 3D avatar identity transfer for outfit personalization. By operating directly on mesh-based Gaussian avatars, our method reuses a user avatar providing head, body shape, and skin tone and a model avatar providing the full-body garment; GSReshape adapts garment retargeting to the Gaussian setting for body-shape alignment, and a two-tier diffusion refinement strategy removes composition and body reshaping artifacts while preserving the pristine outfit and facial details.
 Experiments on THUman2.0 show that AvatarMix achieves higher similarity to source outfits and user identities than state-of-the-art virtual try-on and 3DGS editing baselines, and produces more consistent multi-view results. User studies further confirm that our outputs are preferred in terms of overall realism, view consistency, and facial quality.

{
    \small
    \bibliographystyle{ieeenat_fullname}
    \bibliography{main}
}

\clearpage
\maketitlesupplementary

\begin{table*}[t]
  \centering
  \small
  \caption{\textbf{Comparison of avatar editing paradigms.} We group related work into three paradigms and compare them by inputs, generative model usage, and clothes-body intersection handling. Unlike 2D-to-3D VTON methods that rely on garment inpainting and layered 3D garment approaches that require collision post-processing, AvatarMix composes two Gaussian avatars explicitly and applies diffusion as localized refinement on 3D-consistent rendered images, which improves view consistency and facial identity preservation, and is intersection-free by design.}
  \setlength{\tabcolsep}{5.5pt}
  \label{tab:paradigms}
  \begin{tabular}{p{2.4cm} p{3.2cm} p{3.4cm} p{3.4cm} p{3.0cm}}
    \toprule
    Paradigm & Representative methods & Input conditions & Generative model usage & Clothes--body intersection handling \\
    \midrule
    2D-to-3D VTON &
    VTON360~\cite{vton36025}, GS-VTON~\cite{gsvton24}, GaussianVTON~\cite{gaussianvton24} &
    User multi-view images $+$ garment images &
    Full image generation with garment inpainting & N/A \\
    \midrule
    Layered 3D garments &
    LayGA~\cite{layga24}&
    Two multiview videos & N/A & Post-processing for collision handling \\
    \midrule
    Gaussian avatars 3D composition  &
    AvatarMix (ours) &
    Two Gaussian avatars &
    Local/global refinement on 3D-consistent renderings & N/A \\
    \bottomrule
  \end{tabular}
\end{table*}
\begin{figure*}[t]
  \centering
   \includegraphics[width=\linewidth]{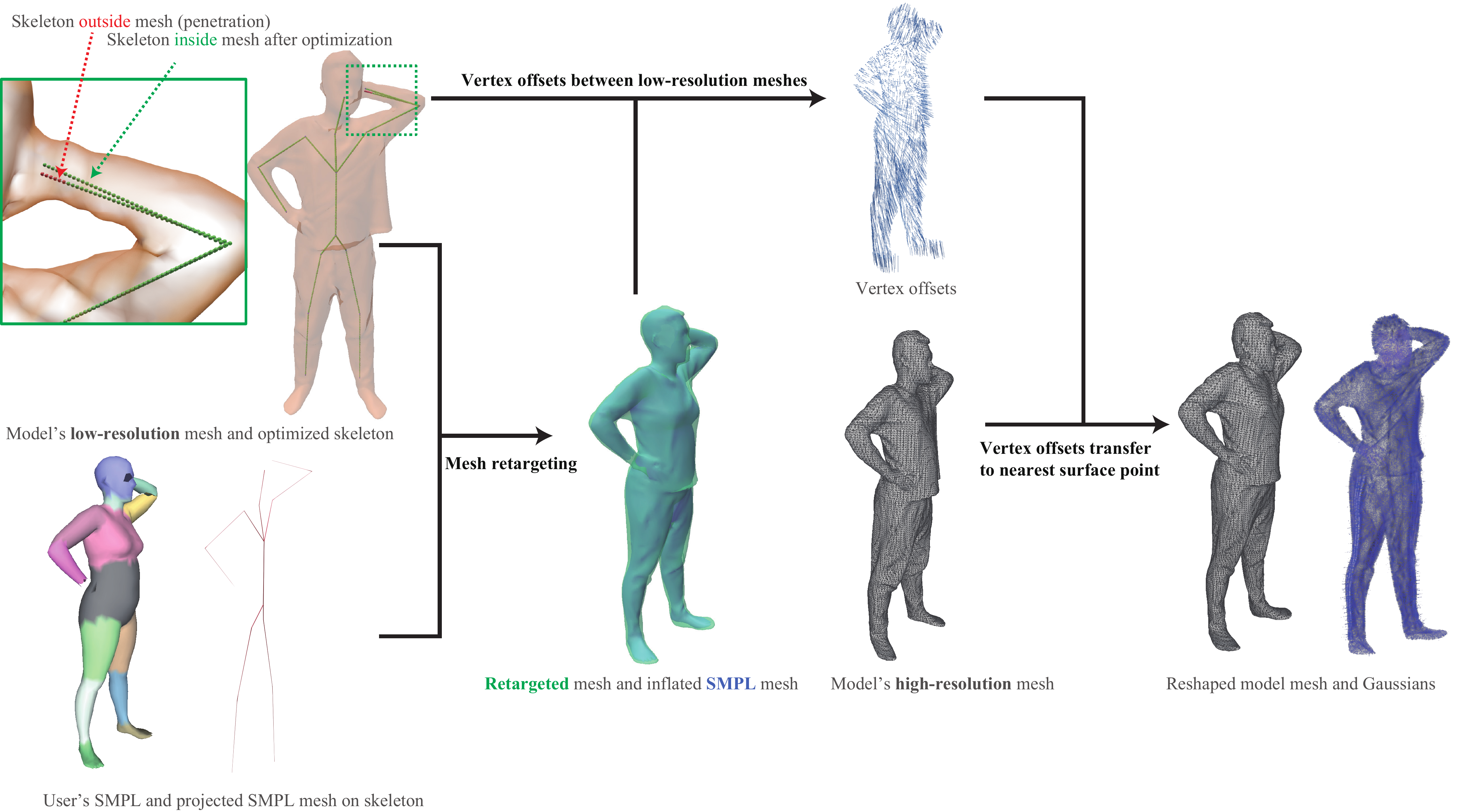}

   \caption{\textbf{GSReshape pipeline overview.} From left to right: starting from the model's low-resolution clothed mesh (top left) and SMPL-X mesh (bottom left), we project the SMPL mesh to skeleton, inflating the SMPL mesh while jointly optimizing the clothed mesh, following the retargeting method of Huang \etal~\cite{clothfit25}. After retargeting, we compute vertex offsets between input and retargeted clothed mesh, and transfer these offsets to the original high-resolution clothed mesh via nearest-surface projection. The Gaussians defined on high-resolution mesh are updated as well.}
   \label{fig:suppl_gsreshape_overview}
\end{figure*}

\begin{figure}[t]
  \centering
  \includegraphics[width=\linewidth]{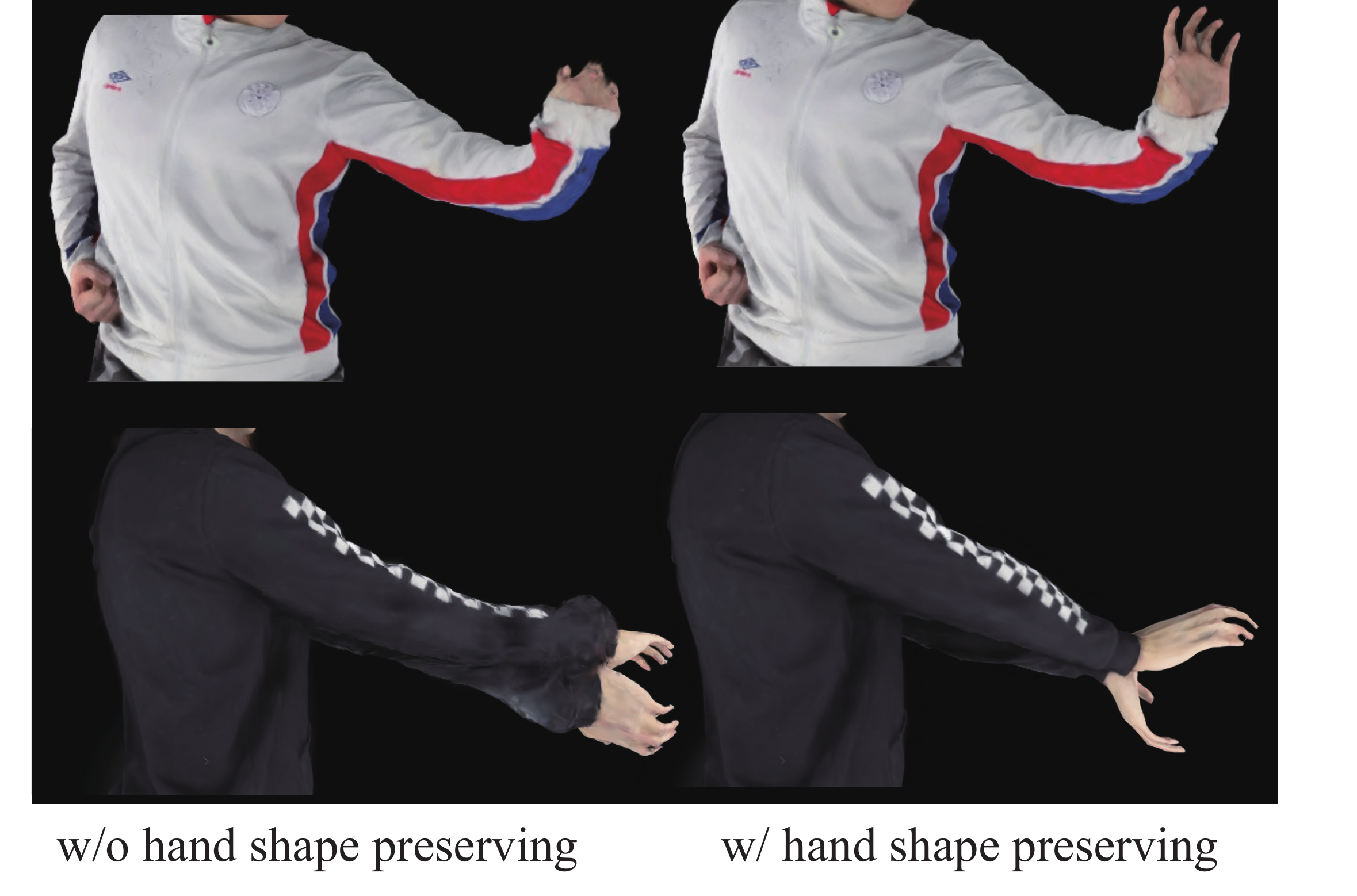}

  \caption{\textbf{Hand-aware skin tightness examples.} First row: high fit weight produces Gaussian artifacts (left) versus our hand shape preserving method (right). Second row: low fit weight creates glove-like hands (left) versus our approach (right). Our semantic weighting strategy achieves better balance between visual fidelity and robustness.}
  \label{fig:suppl_hand}
\end{figure}

\begin{figure}[t]
  \centering
  \includegraphics[width=0.95\linewidth]{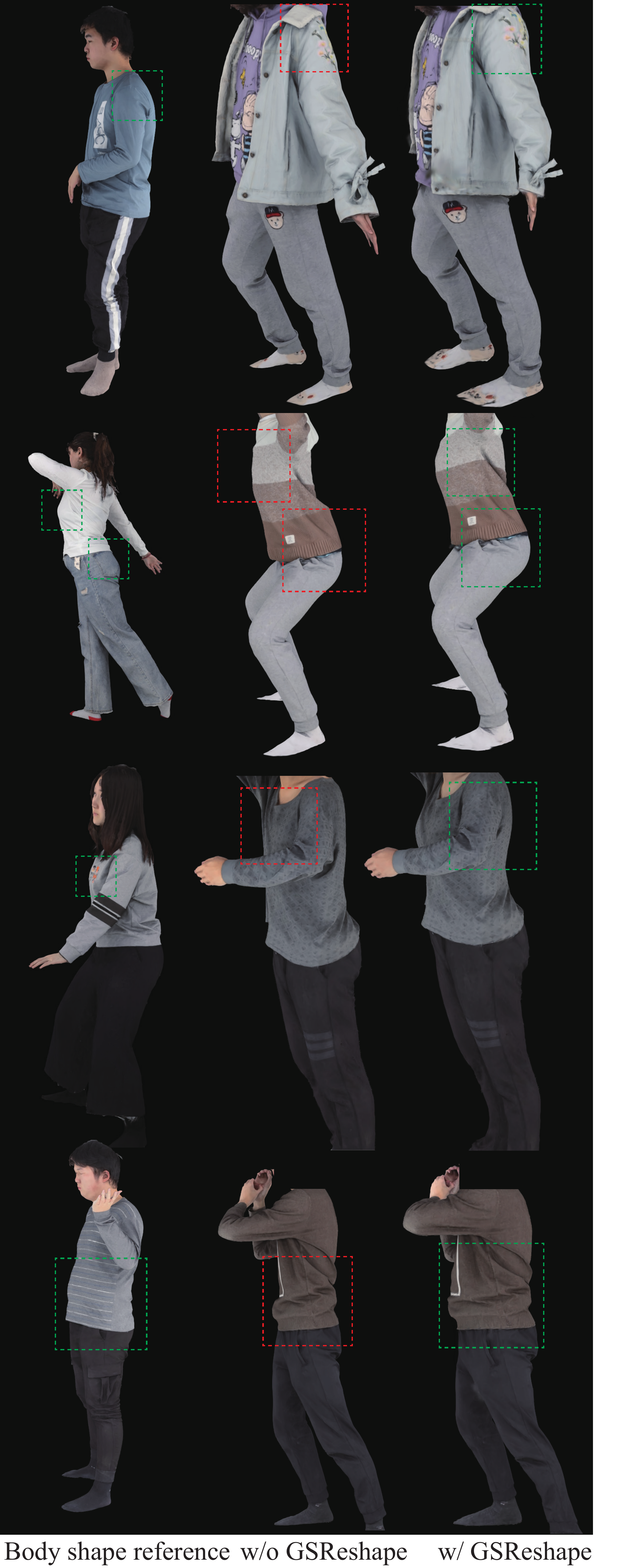}
  \caption{\textbf{Additional ablation on GSReshape.} We visualize the effect of our body reshaping module by comparing the model avatars without GSReshape versus with GSReshape. As shown in the with GSReshape results, the garment adapts smoothly to the user's body shape while preserving details after the body reshaping.}
  \label{fig:suppl_gsreshape}
\end{figure}

\begin{figure*}[t]
  \centering
   \includegraphics[width=0.89\linewidth]{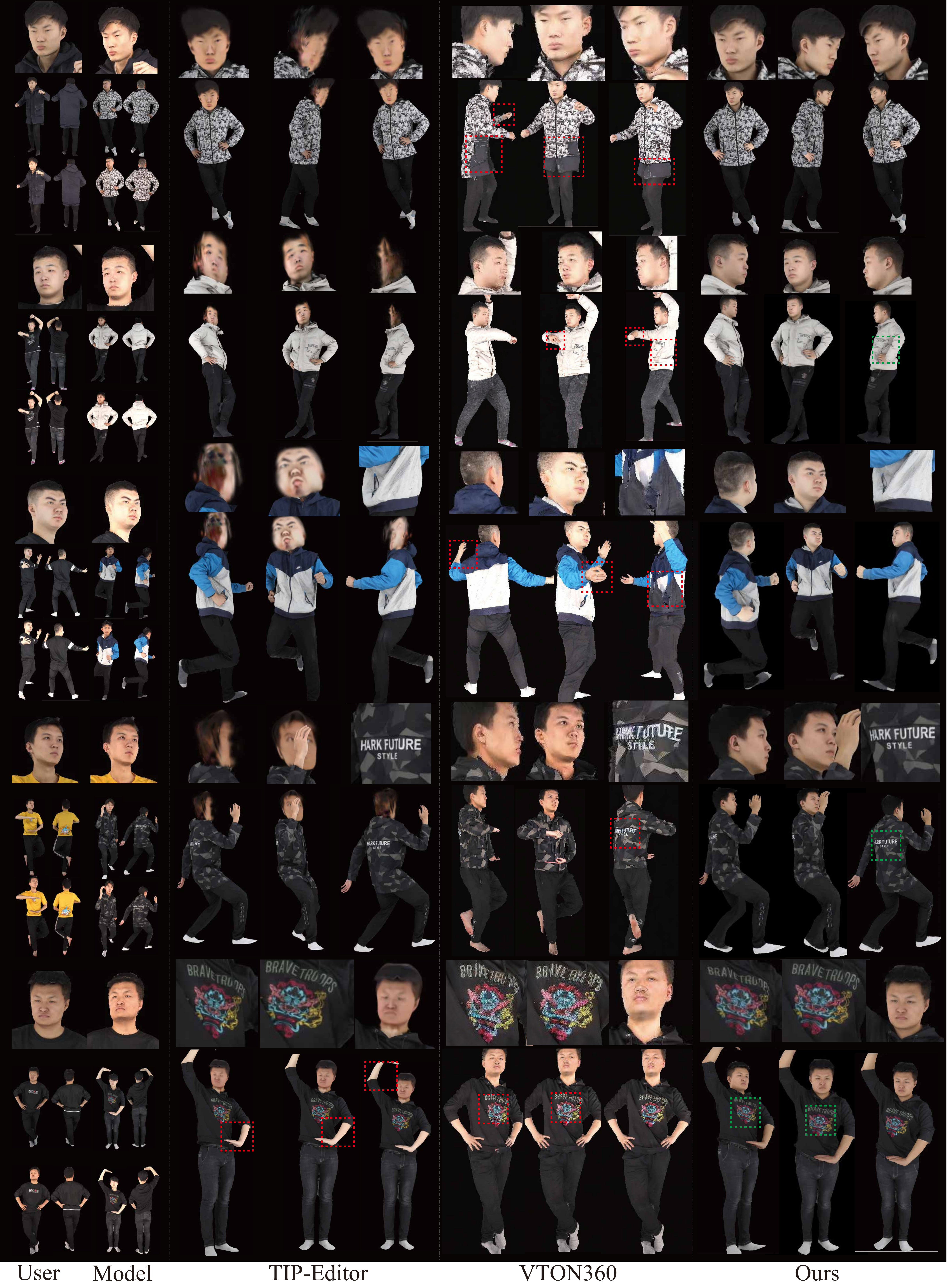}

   \caption{\textbf{Additional comparisons with THUman2.0.} We compare AvatarMix with baselines on more user-model pairs, demonstrating superior preservation of identity and outfit across diverse views.}
   \label{fig:suppl_comparison}
\end{figure*}

\section*{A. Task Setting and Paradigm Comparison}

To clarify the task setting and our position in the design space, we summarize three dominant paradigms for avatar outfit editing and personalization in~\cref{tab:paradigms}: 2D-to-3D virtual try-on, layered 3D garment modeling, and explicit 3D composition on Gaussian avatars. The table contrasts these paradigms by input conditions, use of generative models, and how clothes-body intersections are handled. AvatarMix belongs to the explicit 3D composition paradigm and limits diffusion to localized seam and artifact refinement conditioned on 3D-consistent rendered images, in contrast to artifact-prone garment inpainting in 2D VTON and collision-prone layered 3D garments.

\section*{B. Implementation and Evaluation Details}

\subsection*{B.1 Skin Tone Transfer}

 To harmonize skin appearance between the user's head and the model's body, we operate in Lab color space with opacity-weighted statistics over Gaussian colors. Let $\{\mathbf{c}_i \in [0,1]^3, \alpha_i\}_{i=1}^N$ denote RGB colors and opacities of either the user's facial Gaussians or the model's skin Gaussians, and let $\mathbf{\ell}_i = f_{\text{Lab}}(\mathbf{c}_i)$ be the Lab conversion of the RGB colors. We compute the opacity-weighted mean $\boldsymbol{\mu}$ and variance $\boldsymbol{\sigma}^2$ as
 \begin{equation}
 \boldsymbol{\mu} = \frac{\sum_{i=1}^N \alpha_i \mathbf{\ell}_i}{\sum_{i=1}^N \alpha_i}, \quad
 \boldsymbol{\sigma}^2 = \frac{\sum_{i=1}^N \alpha_i (\mathbf{\ell}_i - \boldsymbol{\mu})^2}{\sum_{i=1}^N \alpha_i}.
 \end{equation}
 We denote the user's facial statistics as $(\boldsymbol{\mu}^{u}, \boldsymbol{\sigma}^{u})$ and the model's skin statistics as $(\boldsymbol{\mu}^{m}, \boldsymbol{\sigma}^{m})$. For each model's skin color $\mathbf{\ell}$ in Lab color space, we perform channel-wise affine transfer
 \begin{equation}
 \mathbf{\ell}' = (\mathbf{\ell} - \boldsymbol{\mu}^{m}) \odot \frac{\boldsymbol{\sigma}^{u}}{\boldsymbol{\sigma}^{m}} + \boldsymbol{\mu}^{u},
\end{equation}
 where $\odot$ denotes element-wise multiplication. The transformed colors $\mathbf{\ell}'$ are then converted back to RGB and assigned to the corresponding skin Gaussians. This global, opacity-aware color transfer matches the model body's skin tone to the user's while preserving local shading and high-frequency detail.

\subsection*{B.2 Diffusion Refinement and GSReshape Implementation}

\noindent\textbf{Diffusion refinement.}
Both SeamFix and FullbodyFix are implemented and trained on top of the pretrained Difix3D+ backbone~\cite{difix3d25}, as described in the main paper. In addition, we attach rank-4 LoRA adapters to the VAE decoder and fine-tune the skip connections between the VAE encoder and decoder following Difix3D. SeamFix is trained for 10 epochs and FullbodyFix for 5 epochs on approximately 19k multi-view double-swapped training samples generated from the THUman2.0 training subjects, using a batch size of 1. For SeamFix, we operate on a cropped square head-and-neck region that is resized to $512\times512$ during both training and testing. At test time we paste the refined crop back into the original image with a feathered blending boundary. For FullbodyFix, we crop a tight bounding box around the full human body, resize this crop to $488\times896$ pixels, and use this resolution during training and inference. All training is conducted on a single NVIDIA RTX A6000 Ada GPU; training SeamFix and FullbodyFix requires roughly 16 and 28 hours, respectively.

\noindent\textbf{GSReshape optimization.}
Our GSReshape module builds on the intersection-free garment retargeting method of Huang \etal~\cite{clothfit25}; an overview of the full pipeline is shown in~\cref{fig:suppl_gsreshape_overview}. Before retargeting, we optimize the SMPL-X skeleton vertices $X = \{x_k\}_{k=1}^{N_s}$ inside a signed distance field $\phi(\cdot)$ of the clothed body mesh in order to remove bone--mesh intersections while preserving bone lengths.
When initializing the skeleton from the SMPL mesh, we attach each vertex to the bone with the largest linear blend skinning (LBS) weight instead of using nearest-distance assignment, which yields a more stable optimization and is visualized by the color-coded SMPL vertices in~\cref{fig:suppl_gsreshape_overview}.
For each bone $(i,j)$, we sample a set of points $\{p_s\}$ along the segment and define an inside penalty
 \begin{equation}
 E_{\text{inside}}(X) = \sum_s \max\bigl(0,\ \phi(p_s) + \delta\bigr)^2,
 \end{equation}
with margin $\delta = 0.1$, a bone-length regularizer
\begin{align}
E_{\text{length}}(X) &= \sum_{(i,j)\in E} \bigl(\|x_i - x_j\|^2 - L_{ij}^2\bigr)^2, \\
L_{ij}^2 &= \|x_i^{(0)} - x_j^{(0)}\|^2,
\end{align}
and an root anchor regularizer
 \begin{equation}
 E_{\text{anchor}}(X) = \|x_r - x_r^{(0)}\|^2.
 \end{equation}
The skeleton optimization objective
\begin{align}
E_{\text{pre}}(X) &= w_{\text{inside}} \, E_{\text{inside}}(X) + w_{\text{len}} \, E_{\text{length}}(X) \\
                  &\quad + w_{\text{anch}} \, E_{\text{anchor}}(X),
\end{align}
with $w_{\text{inside}} = 50.0$, $w_{\text{len}} = 5.0$, and $w_{\text{anch}} = 10.0$, is minimized with the same set of solvers as the method of Huang~\etal, initialized from the original skeleton. We use 40 samples per bone, an SDF voxel size of 0.005, and robust SDF settings (flood-filled sign, hole closing with a 2-voxel radius, and capping of open boundaries), yielding an intersection-free and approximately rigid skeleton used in GSReshape. During the subsequent mesh retargeting, let $V_g = \{v_i\}$ and $X_A = \{x_j\}$ denote garment and avatar vertices. On vertices in the hand region we scale the SDF fit and similarity weights using $m_{\text{fit}}(v_i) = 0.01$ and $m_{\text{sim}}(v_i) = 2.0$, while keeping $m_{\text{fit}}(v_i) = 5, m_{\text{sim}}(v_i) = 1$ elsewhere. We also remove SMPL-X mesh vertices belonging to the hands from the avatar before SDF construction and continuation, which together help avoid glove-like inflation while preserving local hand shape. \cref{fig:suppl_hand} illustrates the effectiveness of this hand-aware design with two examples: the first row compares high fit weight (causing Gaussian artifacts) with our hand shape preserving method, and the second row compares low fit weight (producing glove-like hands) with our approach, demonstrating that our semantic weighting strategy achieves a better balance between visual fidelity and robustness.

\subsection*{B.3 Evaluation Metrics and Protocols}

 We employ three quantitative metrics that capture different aspects of avatar editing quality, complemented by a user study for perceptual evaluation. All metrics are computed on per-view edited images: for VTON360 we use the raw network outputs, for AvatarMix we use the images refined by SeamFix and FullbodyFix, and for TIP-Editor we use the edited rendered images produced by their pipeline.

\noindent\textbf{DINO Similarity for Editing Target.}
To assess how well each method preserves the appearance of the region it edits, we compute DINO~\cite{dinov224} feature similarity, akin to the garment similarity metric used in VTON360. However, since the methods we compare have different editing targets (head and body for our method, upper-body garment for VRON360, and head only for TIP-Editor), we correspondingly adjust the target for computing DINO similarity for fair comparison as follows. For VTON360, whose target is the upper garment only, we compare its edited images against the corresponding front and back garment references using a garment-only segmentation mask. Because VTON360's try-on results may exhibit substantial pose changes relative to the garment images, we restrict this comparison to front ($0^\circ$) and back ($180^\circ$) views, which are the most geometrically aligned and thus conservative in favor of VTON360. For AvatarMix, whose target is the fully clothed body, we compare 36 edited views against renders of the ground-truth model avatar using a clothed-body mask, making the evaluation stricter despite the smaller geometric changes introduced by body reshaping. TIP-Editor performs head-only replacement and leaves the garment and body unchanged in our setting, so this metric is not applicable to TIP-Editor.

\noindent\textbf{Head and Neck DINO Similarity.} To evaluate facial identity preservation and the seamlessness of the neck region, we compute DINO feature similarity on a head-and-neck segmentation mask between the edited images and the ground-truth user avatar. We evaluate this metric over 36 views for all three methods. This protocol is disadvantageous to AvatarMix: after editing, the head and body align with the model's pose, so self-occlusions can differ between the edited and user avatar, which tends to reduce similarity scores even when identity is preserved. Despite this, AvatarMix still achieves the highest head-and-neck DINO similarity.

\noindent\textbf{Warping-based RMSE.} To quantify multi-view consistency, we use a warping-and-RMSE metric computed directly on the edited images instead of the CLIP Direction Consistency Score~\cite{instruct23} used in VTON360. Directional CLIP evaluates whether appearance changes between neighboring views are similar before and after editing, which is suitable when edits mainly affect texture while pose is fixed, as in VTON360's original setting. In our case, both garment appearance and the user's pose can change after editing due to our compositional identity transfer approach; we observe disagreement between Directional CLIP and human judgments of consistency in this setting. We therefore adopt a more direct image-space measure: following the public implementation from the work of Asim \etal~\cite{met3r25}, we first estimate dense 2D correspondences between neighboring views and then measure the root-mean-squared error between one view and the other warped into its coordinate frame. Lower values indicate that details and geometry are stable across viewpoints. We report this metric for all three methods using their respective edited images across all viewpoints.

\section*{C. Limitations and Future Work}

\subsection*{C.1 Limitations}

 While AvatarMix achieves strong results on THUman2.0, several limitations remain. First, our current GSReshape design does not explicitly model detailed hand shape adaptation, which can lead to mismatches at extreme body shape differences. In addition, very loose garments or highly complex accessories may challenge the underlying garment retargeting, occasionally producing wrinkles or folds that differ from the original model.

\subsection*{C.2 Future Work}

One of the future works is
exploring more diverse datasets beyond THUman2.0 to assess generalization across broader clothing styles. Another promising direction is to explore avatar reposing from reconstructed 3D Gaussians (existing reposing works often take a monocular video~\cite{vid2avatar25} or multi-view videos~\cite{animatablegaussians24,layga24} as input), extending user's control on avatar pose after garment personalization.

\section*{D. Additional Qualitative Results}
We provide additional multi-view comparisons on THuman2.0 not shown in the main paper (\cref{fig:suppl_comparison}), demonstrating that AvatarMix maintains facial identity, seamless neck transition, and garment fidelity across challenging poses and lighting conditions compared to VTON360 and TIP-Editor. We also show additional ablation results for GSReshape (\cref{fig:suppl_gsreshape}), illustrating how our retargeting module successfully adapts garments from the model avatar to the user's body shape while preserving garment details. The 360-degree videos of ours and comparison methods, rendered with updated (for ours and TIP-Editor) or reconstructed (for VTON360) Gaussian avatars, can be viewed in the supplementary material and on our project page.

\end{document}